%% file: main.tex
\documentclass[final,5p,times,twocolumn,authoryear]{elsarticle}
\usepackage{float}
\usepackage[T1]{fontenc}
\usepackage{amssymb}
\usepackage{amsmath}
\usepackage{graphicx}
\usepackage{float}
\usepackage{booktabs}
\usepackage[table]{xcolor}
\usepackage{bm}
\usepackage{multirow}
\usepackage{placeins}
\usepackage{cuted}
\usepackage{caption}
\definecolor{stdgray}{gray}{0.48}

\journal{Expert Systems with Applications}

\begin{document}

\begin{frontmatter}

\title{ProSAC-CT: Progressive Spectral--Anatomical Co-Guided Multi-Stage Diffusion Model for Low-Dose CT Denoising}

\author[inst1]{Xuepeng Liu}
\author[inst1]{Zetong Liu}
\author[inst1]{Renyiming Li}
\author[inst1]{Yan Li}
\author[inst3]{Ruiyu Li}
\author[inst2]{Ruili Li\corref{cor1}}
\author[inst2]{Jiayi Ding\corref{cor1}}
\cortext[cor1]{Corresponding authors.\\[-0.5mm]
\begin{tabular}[t]{@{}ll@{}}
\textit{Email addresses:} &
\texttt{li.ruili.t3@dc.tohoku.ac.jp} (Ruili Li),\\
&
\texttt{ding.jiayi.s2@dc.tohoku.ac.jp} (Jiayi Ding)
\end{tabular}}
\author[inst4]{Eichi Takaya}

\affiliation[inst1]{organization={College of Medicine and Biological Information Engineering, Northeastern University},
                   city={Shenyang 110169},
                   country={China}}
\affiliation[inst2]{organization={Tohoku University School of Medicine},
                   city={Sendai 980-8575},
                   country={Japan}}
\affiliation[inst3]{organization={State Key Laboratory of Oncology in South China, Sun Yat-sen University Cancer Center},
                   city={Guangzhou 510060},
                   country={China}}
\affiliation[inst4]{organization={AI Lab, Tohoku University Hospital},
                    city={Sendai},
                    state={Miyagi},
                    country={Japan}}
\begin{abstract}
Low-dose computed tomography (LDCT) reduces radiation exposure but suffers from stronger quantum noise, streak artifacts, and local texture degradation, which can obscure anatomical boundaries and weaken low-contrast structures important for downstream medical image analysis. Diffusion models offer a promising solution for LDCT denoising by progressively recovering normal-dose CT (NDCT) images from degraded LDCT inputs. However, diffusion-based LDCT denoising remains challenging due to insufficient anatomical guidance, uncertain frequency-dependent recovery, and uniform reverse-process modeling, which may blur structural boundaries or over-smooth task-relevant details. We propose ProSAC-CT, a progressive spectral--anatomical co-guided multi-stage diffusion model for image-domain LDCT denoising. ProSAC-CT integrates an anatomical-prior-guided conditioning (APGC) module, a residual frequency-domain decoupling stage (RFDDS), and a time-step-decoupling denoising decoder ($\mathrm{TD}^{3}$). APGC introduces LDCT-derived structural guidance, RFDDS provides frequency-enhanced representations, and $\mathrm{TD}^{3}$ progressively assigns them to different reverse-diffusion stages for anatomical stabilization, boundary refinement, and fine-detail recovery. Experimental results across four LDCT degradation benchmarks show that ProSAC-CT improves image fidelity, structural similarity, perceptual quality, and information preservation over representative methods while better preserving boundary-sensitive anatomical details. Beyond these quantitative gains, downstream anatomical-region classification on Mayo-2020 shows that ProSAC-CT better preserves task-relevant anatomical information than competing denoised images, suggesting that its denoising process suppresses LDCT noise while retaining structures needed for subsequent analysis, which provides a practical denoising solution for low-dose CT applications that aim to reduce patient radiation exposure.  
\end{abstract}

\begin{keyword}
Low-dose CT denoising \sep Diffusion model \sep Spectral modeling \sep Anatomical prior 
\end{keyword}

\end{frontmatter}

\input{sec/intro}

\input{sec/method}

\input{sec/expriment}

\FloatBarrier
\bibliographystyle{elsarticle-harv}
\bibliography{ref}

\end{document}

%% file: sec/intro.tex
\section{Introduction}
Computed tomography (CT) is indispensable in clinical diagnosis and screening. However, normal-dose CT (NDCT) increases cumulative radiation exposure, particularly in oncology follow-up and screening. Therefore, low-dose CT (LDCT) is increasingly preferred to reduce radiation burden. Nevertheless, LDCT protocols inevitably introduce stronger quantum noise, streak artifacts, and degraded texture appearance, which may obscure low-contrast lesions and anatomical boundaries \citep{goldman2007ctdose,smith2009radiation,sodickson2009recurrent,rodrigues2019microcoil}. Consequently, restoring diagnostic image quality from LDCT while preserving CT intensity consistency remains a central challenge in CT reconstruction and post-processing \citep{luo2015lowdose,huynh2016radiomics,kaur2023review}.

From an engineering application perspective, LDCT denoising is not only an image enhancement problem but also a clinically relevant post-processing task for practical CT imaging workflows. In routine screening, follow-up imaging, and dose-sensitive patient monitoring, a denoising model can be deployed after low-dose acquisition and image reconstruction to improve the readability of LDCT images while retaining anatomy-related information for subsequent interpretation. Such a module is particularly useful when raw projection data or scanner-specific reconstruction parameters are unavailable, since image-domain denoising can be integrated into existing picture archiving and communication systems (PACS) or downstream computer-aided diagnosis pipelines. Therefore, an effective LDCT denoising model should not only improve image-level fidelity, but also preserve task-relevant anatomical cues that support downstream analysis.

Early LDCT denoising methods mainly relied on model-based or hand-crafted priors, such as projection-domain filtering, iterative reconstruction, dictionary learning, sparse representation, and non-local regularization \citep{balda2012ray,manduca2009projection,airnet2020,niu2014sparse,kang2022ldctwavelet,yan2023convdict,liu2012adaptive,xu2012dictionary,zheng2018pwls}. Although these methods are interpretable, they often require careful parameter tuning and may struggle to suppress spatially heterogeneous noise while preserving subtle anatomical structures. With the development of deep learning, various neural network architectures have been widely explored for LDCT denoising. CNN-based methods, such as RED-CNN, directly learn the mapping from LDCT to NDCT, achieving efficient noise suppression with compact network designs \citep{redcnn}. Transformer-based architectures further improve LDCT denoising by capturing long-range dependencies and global anatomical context, which facilitates the recovery of structural and textural information \citep{ctformer,amir}. In addition, adversarial learning has been introduced to encourage denoised images to approach the NDCT distribution, thereby improving perceptual realism and texture recovery \citep{unad}. More recently, diffusion models have provided a promising generative paradigm for LDCT denoising by formulating restoration as a progressive reverse process rather than a single-step deterministic mapping \citep{ddpm}. Conditional diffusion and bridge-based models, including I2SB, ResShift, RDDM, and CoreDiff, have shown strong potential for inverse problems and LDCT denoising by gradually transforming degraded LDCT images toward high-quality NDCT images \citep{I2SB,resshift,rddm,corediff,bbdm,pan20232d}.

Despite these advances, existing diffusion-based LDCT denoising methods still face several limitations. First, the progressive noising and reverse denoising process may weaken fine anatomical boundaries. Without explicit structural guidance, diffusion models can degrade fine anatomical structures, reducing the structural cues needed for downstream medical image analysis. Second, LDCT denoising involves different restoration requirements across frequency bands. Low-frequency components mainly preserve the overall anatomical intensity distribution, middle-frequency components reflect edge transitions and structural boundaries, and high-frequency components contain both fine textures and noise. However, existing diffusion-based methods often do not explicitly distinguish these frequency-dependent characteristics, which may lead to insufficient coordination between anatomical consistency, boundary preservation, and noise removal. Third, existing diffusion-based methods often treat the reverse process as a uniform denoising trajectory, without explicitly modeling the stage-specific roles of different sampling steps. This may weaken the progressive coordination between global structure recovery, boundary refinement, and fine-detail reconstruction.

To address these limitations, we propose ProSAC-CT, a progressive spectral–anatomical co-guided multi-stage diffusion model for LDCT denoising. ProSAC-CT integrates an anatomical prior-guided conditioning (APGC) module, a residual frequency-domain decoupling stage (RFDDS), and a time-step-decoupled denoising decoder ($\mathrm{TD}^{3}$) around a shared encoder. First, APGC operates at the encoder stage, extracting an LDCT-derived anatomical prior and injecting it into the shared encoder so that the resulting latent representation is anatomy-aware while preserving CT intensity consistency. Second, the shared encoder produces a single shared latent representation (SLR), which is replicated into three parallel branches. RFDDS then decouples the replicated SLR in each branch into low-, middle-, and high-frequency bands. These bands are then weighted and modulated to enhance the branch in a residual manner. The weighting and modulation are applied separately to each branch, so that every branch receives its own branch-specific enhancement in the frequency domain, making these differently enhanced representations available for different diffusion stages. Third, the three frequency-enhanced representations are fed to the $\mathrm{TD}^{3}$, which decouples the denoising along the time-step axis of the diffusion process. The motivation is that a diffusion process does not denoise uniformly across time-steps. Early time-steps fix the global anatomical structure, while later time-steps refine fine detail, so denoising every time-step through one shared decoder forces it to handle conflicting objectives. $\mathrm{TD}^{3}$ instead decouples the decoder across time-step ranges and lets each range denoise from its own frequency-enhanced representation, so that the frequency information available at each time-step range is tailored to that part of the denoising process. 

Through extensive experiments on four CT datasets, we compare ProSAC-CT against representative CNN-, GAN-, Transformer-, and diffusion-based methods under a unified evaluation protocol, where it achieves the best overall performance across both quantitative and qualitative comparisons. Going beyond image-quality evaluation, we further conduct a downstream six-class anatomical-region classification experiment on Mayo-2020 using ResNet50, Swin-Tiny, and MambaOut-Tiny to examine whether the denoised images retain task-relevant information. The consistent improvement in F1, balanced accuracy (BAcc), and AUC shows that ProSAC-CT does not merely reduce pixel-level noise. It preserves the structural information that subsequent medical image analysis relies on, which provides a practical denoising solution for low-dose CT applications that aim to reduce patient radiation exposure.  


%% file: sec/method.tex
\section{Methodology}

\subsection{Overview}
\begin{figure*}[!t]
\centering
\includegraphics[width=0.9\textwidth]{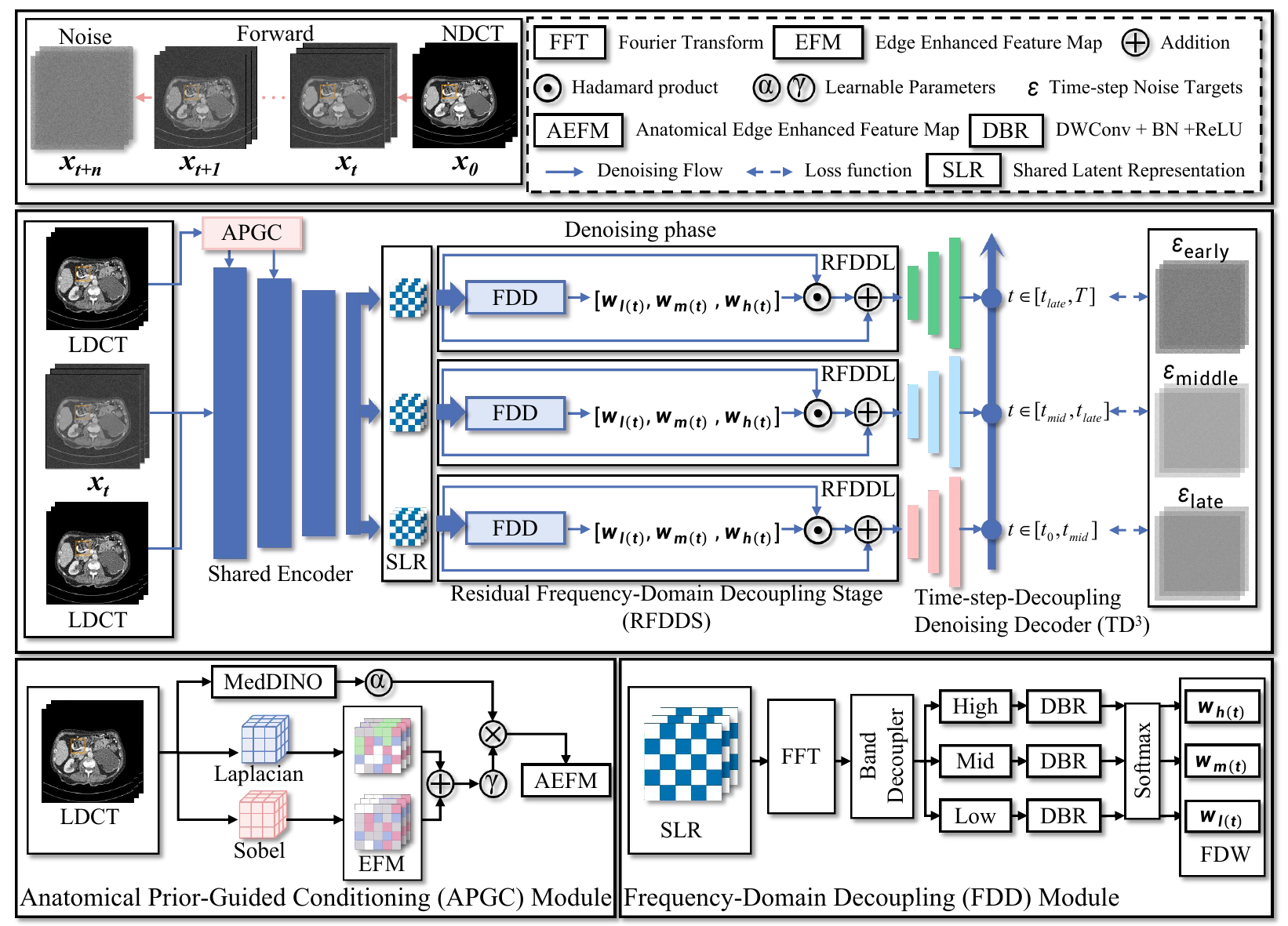}
\caption{Overall architecture of ProSAC-CT for LDCT denoising. The model integrates anatomical-prior-guided conditioning (APGC) module, residual frequency-domain decoupling stage (RFDDS), and time-step-decoupling denoising decoder ($\mathrm{TD}^{3}$) into a conditional diffusion pipeline. APGC module provides LDCT-derived structural guidance, whereas RFDDS separates low-, middle-, and high-frequency components for stage-adaptive residual reweighting, enabling NDCT denoising while preserving CT intensity consistency.}
\label{figmain}
\end{figure*}

Figure~\ref{figmain} illustrates the overall pipeline of ProSAC-CT, which formulates LDCT denoising as an LDCT-conditioned residual diffusion process. Following Residual Denoising Diffusion Model (RDDM)~\citep{rddm}, we define the residual between the LDCT input $x_{\mathrm{LDCT}}$ and its corresponding NDCT reference $x_{\mathrm{NDCT}}$ as
\begin{equation}
r = x_{\mathrm{LDCT}} - x_{\mathrm{NDCT}} .
\end{equation}
Instead of using a standard DDPM forward process that perturbs the clean image with Gaussian noise only, the residual diffusion process jointly introduces residual drift and stochastic noise:
\begin{equation}
x_t =
x_{\mathrm{NDCT}}
+
A_t r
+
B_t \epsilon,
\qquad
\epsilon \sim \mathcal{N}(\mathbf{0},\mathbf{I}),
\end{equation}
where $A_t$ and $B_t$ are the cumulative residual and noise coefficients at time-step $t$, respectively. They are constructed from the residual and noise increment schedules as
\begin{equation}
A_t=\sum_{i=0}^{t}a_i,
\qquad
B_t=\sqrt{\sum_{i=0}^{t}b_i^2}.
\end{equation}

The denoising network jointly predicts the residual and noise components under the LDCT condition:
\begin{equation}
(\hat r_{\theta},\hat\epsilon_{\theta})
=
f_{\theta}(x_t,x_{\mathrm{LDCT}},t).
\end{equation}
The clean NDCT estimate at time-step $t$ is then recovered by removing both predicted components:
\begin{equation}
\hat{x}_{0}
=
x_t
-
A_t\hat r_{\theta}
-
B_t\hat\epsilon_{\theta}.
\end{equation}
During inference, ProSAC-CT adopts a reduced step deterministic residual sampler. For two selected reverse time-steps $t$ and $t'$ with $t'<t$, the update is written as
\begin{equation}
x_{t'}
=
x_t
-
(A_t-A_{t'})\hat r_{\theta}
-
(B_t-B_{t'})\hat\epsilon_{\theta}.
\end{equation}
This residual sampler follows the RDDM denoising principle and avoids the standard DDPM reverse-mean parameterization.

Built upon this backbone, ProSAC-CT incorporates three dedicated components. The APGC module extracts an LDCT-derived anatomical prior and injects it at the encoder stage, making the latent representation anatomy-aware while preserving CT intensity consistency. Then, the encoder produces a shared latent representation (SLR), which is replicated into three parallel branches. The RFDDS module decomposes the replicated SLR in each branch into low-, middle-, and high-frequency bands, and applies branch-specific weighting and modulation to enhance it in a residual manner. Finally, the $\mathrm{TD}^{3}$
decouples the denoising along the time-step axis, letting each time-step range denoise from its own frequency-enhanced representation. We elaborate on each component in the following subsections.

\subsection{Anatomical Prior-Guided Conditioning (APGC) Module}
A diffusion model denoises from the noise distribution without any explicit anatomical constraint, so it tends to blur structural boundaries and introduce artifacts that are not faithful to the underlying anatomy. To provide such a constraint, the APGC module derives an anatomical prior directly from the LDCT input itself, without any external image or auxiliary data. This prior then guides the denoising, keeping the recovered image consistent with the true anatomical structure.

Given the LDCT input $x_{LDCT}\in \mathbb{R}^{C \times H \times W}$, APGC module extracts the prior through three branches. MedDINOv3 \citep{li2025meddinov3}, pretrained on large-scale medical images, provides an anatomical feature map (AFM) that describes the anatomical structure. In parallel, Sobel and Laplacian operators are complementary: Sobel captures the main structural contours and Laplacian supplements the fine boundaries, each producing an edge feature map (EFM):
\begin{equation}
\begin{aligned}
F_{\mathrm{AFM}} &= \mathrm{MedDINOv3}(x_{\mathrm{LDCT}}), \\
F_{\mathrm{EFM}}^{\mathrm{Sob}} &= \mathrm{Sob}(x_{\mathrm{LDCT}}), \qquad
F_{\mathrm{EFM}}^{\mathrm{Lap}} = \mathrm{Lap}(x_{\mathrm{LDCT}}), \\
F_{\mathrm{EFM}} &= F_{\mathrm{EFM}}^{\mathrm{Sob}} + F_{\mathrm{EFM}}^{\mathrm{Lap}}
\end{aligned}
\end{equation}
where $F_{\mathrm{AFM}}, F_{\mathrm{EFM}}^{\mathrm{Sob}}, F_{\mathrm{EFM}}^{\mathrm{Lap}}, F_{\mathrm{EFM}} \in \mathbb{R}^{C \times H \times W}$.

The anatomical feature provides the global structural skeleton, while the two edge maps respectively reinforce the coarse contours and the fine boundaries. We therefore take the AFM as the base and aggregate the two edge cues into a learnable modulation that modulates it through element-wise multiplication, yielding the anatomical-edge enhanced feature map (AEFM):

\begin{equation}
    F_{\mathrm{AEFM}} = (\alpha \odot F_{\mathrm{AFM}}) \odot (\gamma \odot F_{\mathrm{EFM}}),
\end{equation}
where $\alpha, \gamma \in \mathbb{R}^{C}$ are learnable channel-wise weights broadcast across the spatial dimensions, and $\odot$ denotes element-wise multiplication. The aggregated edge map $F_{\mathrm{EFM}}$, learnably modulated by $\gamma$, forms an edge-aware modulation map that reweights the anatomical feature $F_{\mathrm{AFM}}$ scaled by $\alpha$, yielding the anatomy-edge fused map $F_{\mathrm{AEFM}}$.

The fused map $F_{\mathrm{AEFM}}$ is injected into the shared encoder as the anatomical prior, conditioning the encoding so that the resulting latent representation is anatomy-aware. Finally, the shared encoder produces the SLR, denoted $F_{\mathrm{SLR}} \in \mathbb{R}^{C \times H \times W}$:
\begin{equation}
    F_{\mathrm{SLR}} = \mathrm{Encoder}(x_{\mathrm{LDCT}}, F_{\mathrm{AEFM}}),
\end{equation}
which is then processed by RFDDS for frequency-domain enhancement.

\subsection{Residual Frequency-Domain Decoupling Stage (RFDDS)}
Standard diffusion-based denoising generally processes feature representations through a unified reverse path, without explicitly modeling the distinct roles of different frequency components. However, in LDCT denoising, low-frequency components are closely related to global anatomical consistency, middle-frequency components mainly capture structural transitions and boundaries, and high-frequency components contain fine details as well as noise-sensitive responses. Treating these components in an entangled manner may weaken the coordination between anatomical preservation, boundary refinement, and noise suppression. Motivated by this, we design RFDDS to introduce explicit frequency-aware modulation into the latent representation. Specifically, RFDDS decomposes each branch into low-, middle-, and high-frequency components, estimates band-wise channel weights, and residually enhances the original branch with the reweighted frequency components.

RFDDS consists of three parallel residual frequency-domain decoupling layers (RFDDL). Since the three layers are structurally identical, we describe the operation of a single RFDDL. Given the SLR, denoted $F_{\mathrm{SLR}} \in \mathbb{R}^{C \times H \times W}$, RFDDL employs a frequency-domain decoupling (FDD) module to produce the band-wise frequency-domain weights (FDW).

Within the FDD, the branch is first transformed into the frequency domain and shifted to the centered Fourier domain:
\begin{equation}
    F^{\mathcal{F}}_{c}
    =
    \operatorname{fftshift}
    \left(
    \mathcal{F}(F_{\mathrm{SLR}})
    \right),
\end{equation}
where $\mathcal{F}(\cdot)$ denotes the two-dimensional FFT, and $\operatorname{fftshift}(\cdot)$ moves the zero-frequency component to the center of the spectrum.

Let $(u,v)$ denote the frequency coordinates and $(u_0,v_0)$ denote the center of the frequency grid. The normalized radial frequency magnitude is defined as
\begin{equation}
    r(u,v)
    =
    \frac{
    \sqrt{(u-u_0)^2+(v-v_0)^2}
    }{
    \sqrt{(H/2)^2+(W/2)^2}
    }.
\end{equation}
The low-, middle-, and high-frequency masks are defined in the centered Fourier domain as
\begin{equation}
    \mathcal{M}_{l}(u,v)
    =
    \mathbf{1}\left[r(u,v)\le r_{1}\right],
\end{equation}
\begin{equation}
    \mathcal{M}_{m}(u,v)
    =
    \mathbf{1}\left[r_{1}<r(u,v)\le r_{2}\right],
\end{equation}
\begin{equation}
    \mathcal{M}_{h}(u,v)
    =
    \mathbf{1}\left[r(u,v)>r_{2}\right],
\end{equation}
where $0<r_{1}<r_{2}<1$ are fixed cutoff thresholds used for all experiments. The masks $\mathcal{M}_{l}$, $\mathcal{M}_{m}$, and $\mathcal{M}_{h}\in\{0,1\}^{H\times W}$ are shared across all channels.

Each frequency component is obtained by applying the corresponding mask in the centered Fourier domain and then transforming it back to the spatial domain:
\begin{equation}
    F_b =
    \operatorname{Re}
    \left[
    \mathcal{F}^{-1}
    \left(
    \operatorname{ifftshift}
    \left(
    \mathcal{M}_{b} \odot F^{\mathcal{F}}_{c}
    \right)
    \right)
    \right],
    \quad b\in\{l,m,h\},
\end{equation}
where $\operatorname{ifftshift}(\cdot)$ restores the original FFT ordering before inverse transformation, $\mathcal{F}^{-1}(\cdot)$ denotes the inverse FFT, $\operatorname{Re}[\cdot]$ extracts the real-valued spatial response, and $\odot$ denotes element-wise multiplication. 
To estimate the importance of each frequency band, each component is processed by a DBR block and then compressed by global average pooling to obtain a channel descriptor:
\begin{equation}
    \mathbf{z}_{b} = \mathrm{GAP}\big(\mathrm{DBR}(F_{b})\big) \in \mathbb{R}^{C},
    \quad b \in \{l, m, h\}.
\end{equation}
Here, DBR denotes the DWConv--BN--ReLU unit. In our implementation, it consists of a $3\times3$ depthwise convolution with stride $1$ and padding $1$, followed by batch normalization and ReLU activation. It is applied independently to each decoupled frequency component to extract lightweight band-specific responses before global average pooling.

The three band descriptors are then stacked along the frequency-band dimension and normalized by softmax to produce channel-wise FDW:
\begin{equation}
    [\,\mathbf{w}_{l},\,\mathbf{w}_{m},\,\mathbf{w}_{h}\,]
    =
    \mathrm{Softmax}
    \left(
    [\,\mathbf{z}_{l},\,\mathbf{z}_{m},\,\mathbf{z}_{h}\,]
    \right),
\end{equation}
where $\mathbf{w}_{l}, \mathbf{w}_{m}, \mathbf{w}_{h} \in \mathbb{R}^{C}$ denote the channel-wise FDW for the low-, middle-, and high-frequency components, respectively. The softmax operation is applied across the three frequency bands for each channel, so that the weights indicate the relative contribution of different frequency components within the same channel. Finally, the reweighted frequency components are aggregated as a frequency-aware residual response and added back to the original branch with a learnable channel-wise scaling vector:
\begin{equation}
    \hat{F}
    =
    F_{\mathrm{SLR}}
    +
    \boldsymbol{\eta}
    \odot
    \sum_{b \in \{l,m,h\}}
    \mathbf{w}_{b} \odot F_{b},
\end{equation}
where $\boldsymbol{\eta} \in \mathbb{R}^{C}$ is a learnable channel-wise scaling vector that controls the contribution of the frequency-aware residual response. Both $\boldsymbol{\eta}$ and $\mathbf{w}_{b}$ are broadcast along the spatial dimensions for channel-wise reweighting. $\odot$ denotes element-wise multiplication, and $\hat{F} \in \mathbb{R}^{C \times H \times W}$ is the frequency-enhanced representation of this branch.
\subsection{Time-step-Decoupling Denoising Decoder ($\mathrm{TD}^{3}$) }
Existing diffusion-based denoising methods often treat the reverse process as a uniform denoising trajectory, without explicitly modeling the stage-specific roles of different sampling steps. However, the denoising requirements vary across the reverse diffusion process: high-noise time-steps mainly require global structure stabilization, intermediate time-steps focus more on structural and boundary refinement, and low-noise time-steps emphasize fine-detail consolidation and residual noise suppression. Using a single undifferentiated decoder for all time-steps may entangle these stage-specific denoising objectives and make the role of each denoising stage less explicit. Motivated by this, we propose $\mathrm{TD}^{3}$, which partitions the reverse diffusion trajectory into three noise regimes and assigns stage-specific denoising branches to better match the denoising demands at different time-steps.

As described in the overview, the APGC-enhanced encoder produces a single shared latent representation, denoted as $F_{\mathrm{SLR}} \in \mathbb{R}^{C \times H \times W}$. 
To provide stage-specific frequency guidance for $\mathrm{TD}^{3}$, $F_{\mathrm{SLR}}$ is replicated into three parallel streams, and each stream is processed by an independent RFDDS branch. 
This design allows all denoising stages to share the same anatomy-aware latent basis while receiving different frequency-enhanced representations:
\begin{equation}
    \hat{F}_{s(t)}
    =
    \mathrm{RFDDS}_{s(t)}
    \left(
    F_{\mathrm{SLR}}
    \right),
    \quad
    {s(t)} \in \{\mathrm{early}, \mathrm{middle}, \mathrm{late}\}
\end{equation}
where $\hat{F}_{\mathrm{early}}$, $\hat{F}_{\mathrm{middle}}$, and $\hat{F}_{\mathrm{late}} \in \mathbb{R}^{C \times H \times W}$ denote the frequency-enhanced representations generated from the same shared latent representation but independently modulated for different denoising stages. 
For a given time-step $t$, $\mathrm{TD}^{3}$ selects the corresponding representation $\hat{F}_{s(t)}$ according to the stage indicator $s(t)$.

Then, $\mathrm{TD}^{3}$ partitions the reverse diffusion trajectory into three denoising stages according to the noise schedule. Since reverse denoising proceeds from $T$ to $0$, the stage indicator is defined as
\begin{equation}
s(t)=
\begin{cases}
\mathrm{early}, & t \in (t_{\mathrm{late}}, T], \\
\mathrm{middle}, & t \in (t_{\mathrm{mid}}, t_{\mathrm{late}}], \\
\mathrm{late}, & t \in [0, t_{\mathrm{mid}}].
\end{cases}
\end{equation}
where $t_{\mathrm{late}}=0.7T$ and $t_{\mathrm{mid}}=0.3T$ are predefined thresholds that divide the 1,000-step reverse diffusion process into high-, intermediate-, and low-noise regimes.

Accordingly, for a given time-step $t$, $\mathrm{TD}^{3}$ first determines the current denoising stage through $s(t)$ and then uses the corresponding RFDDS-enhanced representation $\hat{F}_{s(t)}$ for stage-specific residual-noise prediction. Different from the standard DDPM reverse-mean parameterization, $\mathrm{TD}^{3}$ follows the residual-noise formulation of RDDM:
\begin{equation}
\left(
\hat r_{\theta}^{s(t)},
\hat\epsilon_{\theta}^{s(t)}
\right)
=
f_{\theta}^{s(t)}
\left(
x_t,
x_{\mathrm{LDCT}},
t,
\hat F_{s(t)}
\right).
\end{equation}
The clean NDCT estimate is then recovered as
\begin{equation}
\hat{x}_{0}^{s(t)}
=
x_t
-
A_t\hat r_{\theta}^{s(t)}
-
B_t\hat\epsilon_{\theta}^{s(t)}.
\end{equation}
During inference, for two selected reverse time-steps $t$ and $t'$ with $t'<t$, the stage-aware deterministic residual sampling update is written as
\begin{equation}
x_{t'}
=
x_t
-
(A_t-A_{t'})
\hat r_{\theta}^{s(t)}
-
(B_t-B_{t'})
\hat\epsilon_{\theta}^{s(t)}.
\end{equation}
where $\hat r_{\theta}^{s(t)}$ and $\hat\epsilon_{\theta}^{s(t)}$ denote the residual and noise predictions produced by the denoising branch corresponding to the current time-step stage. $\hat F_{s(t)}$ denotes the RFDDS-enhanced representation selected for the corresponding denoising stage.

We then define a stage-dependent objective. For a time-step $t$ assigned to stage $s(t)$, the training loss jointly supervises residual prediction, noise prediction, and image-domain reconstruction:
\begin{equation}
\begin{aligned}
\mathcal{L}_{s(t)}
=&\;
\lambda^{s(t)}_{r}
\left\|
r-\hat r_{\theta}^{s(t)}
\right\|_{2}^{2}
+
\lambda^{s(t)}_{\epsilon}
\left\|
\epsilon-\hat\epsilon_{\theta}^{s(t)}
\right\|_{2}^{2}
\\
&+
\lambda^{s(t)}_{x}
\left\|
\hat{x}_{0}^{s(t)}-x_{\mathrm{NDCT}}
\right\|_{2}^{2}.
\end{aligned}
\end{equation}
The stage-specific loss is selected according to the time-step regime:
\begin{equation}
\mathcal{L}_{\mathrm{TD}^{3}}(t)
=
\begin{cases}
\mathcal{L}_{\mathrm{early}}, 
& t \in (t_{\mathrm{late}}, T],\\
\mathcal{L}_{\mathrm{middle}}, 
& t \in (t_{\mathrm{mid}}, t_{\mathrm{late}}],\\
\mathcal{L}_{\mathrm{late}}, 
& t \in [0,t_{\mathrm{mid}}].
\end{cases}
\end{equation}
where $r=x_{\mathrm{LDCT}}-x_{\mathrm{NDCT}}$ is the residual target, $\epsilon\sim\mathcal{N}(\mathbf{0},\mathbf{I})$ is the sampled noise, and $\hat{x}_{0}^{s(t)}$ is the stage-specific clean estimate recovered from residual-noise prediction. The coefficients $\lambda^{s(t)}_{r}$, $\lambda^{s(t)}_{\epsilon}$, and $\lambda^{s(t)}_{x}$ balance residual prediction, noise prediction, and image-domain reconstruction for the corresponding denoising stage.

%% file: sec/expriment.tex
\section{Experiment and Results}

\subsection{Datasets and Implementation Details}
\label{sec:datasets_implementation}

\textbf{Datasets and compared methods.}
We evaluate the proposed model on four LDCT denoising degradation benchmarks, including two clinical paired CT datasets and two synthetic degradation datasets. Specifically, the NIH-AAPM-Mayo Clinic Low Dose CT Grand Challenge dataset (Mayo-2016) \citep{mayo2016}, and the Mayo Low Dose CT Image and Projection Data collection (Mayo-2020) \citep{mayo2020} are used for clinical low-dose denoising. In addition, the QIN-LUNG-CT dataset (QIN-Lung) \citep{qinlung} and the LoDoPaB dataset \citep{lodopab} are used to evaluate robustness under projection-domain simulation and reconstruction-based synthetic degradation settings.

To evaluate the performance of the proposed ProSAC-CT, we select representative methods from three categories for comparison. 
First, we use RDDM~\citep{rddm} as the baseline diffusion model, since ProSAC-CT is built upon a residual diffusion formulation. 
Second, we compare with state-of-the-art LDCT denoising methods, including the CNN-based RED-CNN~\citep{redcnn}, anatomy-aware UNAD~\citep{unad}, GAN-based DU-GAN~\citep{dugan}, Transformer-based CTformer~\citep{ctformer}, and diffusion-based CoreDiff~\citep{corediff}. 
Third, we include representative general image denoising and diffusion-based denoising methods, including AMIR~\citep{amir}, I2SB~\citep{I2SB}, and ResShift~\citep{resshift}. 
AMIR is a universal medical image restoration framework with task-adaptive routing, while I2SB and ResShift are diffusion/bridge-based denoising models. 
All methods are evaluated under identical train/test splits, preprocessing pipelines, and a unified three-slice input protocol when applicable.
\begin{table*}[!t]
\centering
\caption{Quantitative comparison across four CT degradation benchmarks. Values are reported as mean with standard deviation. PSNR is measured in dB, SSIM is reported as a percentage, and higher values indicate better performance for all listed metrics. Best results are highlighted in bold.}
\label{tab:sota_all_ct}
\scriptsize
\setlength{\tabcolsep}{3pt}
\renewcommand{\arraystretch}{1.0}
\newcommand{\stdt}[1]{{\fontsize{5}{6}\selectfont$\pm$#1}}
\begin{tabular*}{\textwidth}{@{\extracolsep{\fill}}lccccc}
\toprule
\textbf{Method} & \textbf{PSNR $\uparrow$} & \textbf{SSIM (\%) $\uparrow$} & \textbf{FSIM $\uparrow$} & \textbf{NQM $\uparrow$} & \textbf{VIF $\uparrow$} \\
\midrule
\multicolumn{6}{@{}l}{\textbf{Mayo-2016}} \\
\midrule
RED-CNN & 29.1764\stdt{1.7698} & 85.90\stdt{0.19} & 0.9421\stdt{0.0016} & 26.4972\stdt{0.1737} & 0.4468\stdt{0.0046} \\
UNAD & 29.3101\stdt{1.7990} & 86.10\stdt{0.10} & 0.9419\stdt{0.0016} & 26.7313\stdt{0.1775} & 0.4516\stdt{0.0047} \\
DU-GAN & 28.5081\stdt{1.8903} & 86.24\stdt{0.18} & 0.9500\stdt{0.0013} & 25.8911\stdt{0.1824} & 0.4329\stdt{0.0049} \\
CTformer & 28.1315\stdt{1.6408} & 83.40\stdt{0.13} & 0.9366\stdt{0.0012} & 25.7059\stdt{0.1907} & 0.3982\stdt{0.0043} \\
AMIR & 29.3829\stdt{1.8117} & 86.27\stdt{0.10} & 0.9423\stdt{0.0017} & 26.8095\stdt{0.1792} & 0.4537\stdt{0.0048} \\
I2SB & 28.4060\stdt{1.8391} & 84.37\stdt{0.14} & 0.9443\stdt{0.0014} & 25.8141\stdt{0.1799} & 0.4214\stdt{0.0048} \\
ResShift & 27.5272\stdt{1.6399} & 82.46\stdt{0.16} & 0.9392\stdt{0.0014} & 25.2101\stdt{0.1667} & 0.3842\stdt{0.0041} \\
RDDM & 27.7347\stdt{1.8401} & 83.44\stdt{0.15} & 0.9421\stdt{0.0014} & 24.9625\stdt{0.1850} & 0.4086\stdt{0.0047} \\
CoreDiff & 28.5388\stdt{1.7503} & 85.28\stdt{0.19} & 0.9421\stdt{0.0015} & 25.1415\stdt{0.1700} & 0.4374\stdt{0.0047} \\
\textbf{Ours} & \textbf{31.0847}\stdt{1.7026} & \textbf{88.46}\stdt{0.12} & \textbf{0.9587}\stdt{0.0011} & \textbf{28.4132}\stdt{0.1522} & \textbf{0.4986}\stdt{0.0042} \\
\midrule
\multicolumn{6}{@{}l}{\textbf{Mayo-2020}} \\
\midrule
RED-CNN & 26.8520\stdt{1.5413} & 84.27\stdt{0.19} & 0.9334\stdt{0.0027} & 26.3699\stdt{0.3497} & 0.3928\stdt{0.0113} \\
UNAD & 27.1626\stdt{1.8297} & 84.78\stdt{0.10} & 0.9369\stdt{0.0027} & 27.2367\stdt{0.3644} & 0.4061\stdt{0.0121} \\
DU-GAN & 26.0511\stdt{1.9162} & 84.47\stdt{0.12} & 0.9452\stdt{0.0019} & 26.2336\stdt{0.3850} & 0.3868\stdt{0.0119} \\
CTformer & 25.2439\stdt{1.2615} & 79.94\stdt{0.12} & 0.9318\stdt{0.0024} & 24.3916\stdt{0.4201} & 0.3321\stdt{0.0115} \\
AMIR & 26.8019\stdt{1.8425} & 84.69\stdt{0.10} & 0.9448\stdt{0.0019} & 27.2653\stdt{0.3452} & 0.3995\stdt{0.0119} \\
I2SB & 26.3810\stdt{1.0984} & 83.70\stdt{0.13} & 0.9447\stdt{0.0019} & 26.8764\stdt{0.3506} & 0.3898\stdt{0.0122} \\
ResShift & 25.4985\stdt{1.6851} & 81.72\stdt{0.17} & 0.9384\stdt{0.0020} & 25.5959\stdt{0.3157} & 0.3557\stdt{0.0105} \\
RDDM & 26.1135\stdt{1.2207} & 83.65\stdt{0.11} & 0.9440\stdt{0.0019} & 26.5448\stdt{0.3579} & 0.3834\stdt{0.0125} \\
CoreDiff & 26.4809\stdt{1.9507} & 84.53\stdt{0.19} & 0.9441\stdt{0.0019} & 27.0110\stdt{0.3337} & 0.3922\stdt{0.0121} \\
\textbf{Ours} & \textbf{28.9435}\stdt{1.2179} & \textbf{87.19}\stdt{0.13} & \textbf{0.9564}\stdt{0.0018} & \textbf{29.0076}\stdt{0.3109} & \textbf{0.4538}\stdt{0.0107} \\
\midrule
\multicolumn{6}{@{}l}{\textbf{LoDoPaB}} \\
\midrule
RED-CNN & 39.5174\stdt{1.9825} & 92.72\stdt{0.19} & 0.9380\stdt{0.0051} & 25.1918\stdt{0.4501} & 0.5497\stdt{0.0171} \\
UNAD & 39.8474\stdt{1.0994} & 93.17\stdt{0.19} & 0.9435\stdt{0.0048} & 25.9575\stdt{0.4607} & 0.5758\stdt{0.0177} \\
DU-GAN & 38.3323\stdt{1.7391} & 90.89\stdt{0.12} & 0.9461\stdt{0.0029} & 24.1649\stdt{0.4616} & 0.5220\stdt{0.0168} \\
CTformer & 35.1568\stdt{1.8402} & 89.51\stdt{0.10} & 0.9161\stdt{0.0035} & 19.2343\stdt{0.3552} & 0.4296\stdt{0.0134} \\
AMIR & 40.2329\stdt{1.3173} & 93.52\stdt{0.10} & 0.9392\stdt{0.0054} & 26.2176\stdt{0.4712} & 0.5810\stdt{0.0177} \\
I2SB & 40.6003\stdt{1.6917} & 92.87\stdt{0.13} & 0.9542\stdt{0.0035} & 27.2775\stdt{0.4969} & 0.5780\stdt{0.0181} \\
ResShift & 37.9829\stdt{1.5375} & 91.68\stdt{0.17} & 0.9497\stdt{0.0033} & 24.1955\stdt{0.4317} & 0.5387\stdt{0.0173} \\
RDDM & 39.8229\stdt{1.1982} & 92.95\stdt{0.11} & 0.9468\stdt{0.0040} & 26.8205\stdt{0.4828} & 0.5684\stdt{0.0176} \\
CoreDiff & 39.2431\stdt{1.2441} & 92.68\stdt{0.16} & 0.9410\stdt{0.0056} & 25.0273\stdt{0.4932} & 0.5810\stdt{0.0177} \\
\textbf{Ours} & \textbf{42.3176}\stdt{1.2845} & \textbf{95.41}\stdt{0.17} & \textbf{0.9667}\stdt{0.0028} & \textbf{29.0248}\stdt{0.4285} & \textbf{0.6325}\stdt{0.0163} \\
\midrule
\multicolumn{6}{@{}l}{\textbf{QIN-Lung}} \\
\midrule
RED-CNN & 32.6955\stdt{1.1305} & 89.45\stdt{0.10} & 0.8372\stdt{0.0040} & 29.3961\stdt{0.3471} & 0.1615\stdt{0.0053} \\
UNAD & 32.8567\stdt{1.1943} & 89.52\stdt{0.19} & 0.9659\stdt{0.0015} & 31.1246\stdt{0.2756} & 0.4744\stdt{0.0073} \\
DU-GAN & 31.5011\stdt{1.2008} & 87.31\stdt{0.15} & 0.9666\stdt{0.0011} & 30.4175\stdt{0.2818} & 0.4412\stdt{0.0070} \\
CTformer & 30.6788\stdt{1.7327} & 86.67\stdt{0.11} & 0.8355\stdt{0.0041} & 29.4065\stdt{0.3355} & 0.1519\stdt{0.0049} \\
AMIR & 33.1691\stdt{1.3509} & 89.86\stdt{0.10} & 0.9648\stdt{0.0017} & 31.3993\stdt{0.2836} & 0.4854\stdt{0.0078} \\
I2SB & 32.2886\stdt{1.3027} & 88.39\stdt{0.14} & 0.9666\stdt{0.0012} & 30.2708\stdt{0.2792} & 0.4601\stdt{0.0076} \\
ResShift & 31.4410\stdt{1.0942} & 87.47\stdt{0.17} & 0.9635\stdt{0.0012} & 29.2861\stdt{0.2680} & 0.4351\stdt{0.0071} \\
RDDM & 32.0167\stdt{1.1728} & 88.35\stdt{0.14} & 0.9633\stdt{0.0014} & 29.7522\stdt{0.2812} & 0.4482\stdt{0.0072} \\
CoreDiff & 32.9534\stdt{1.2357} & 89.71\stdt{0.19} & 0.9664\stdt{0.0013} & 30.8759\stdt{0.2793} & 0.4799\stdt{0.0074} \\
\textbf{Ours} & \textbf{34.8468}\stdt{1.0862} & \textbf{92.37}\stdt{0.13} & \textbf{0.9752}\stdt{0.0010} & \textbf{33.0127}\stdt{0.2461} & \textbf{0.5318}\stdt{0.0065} \\
\bottomrule
\end{tabular*}
\end{table*}

\textbf{Mayo-2016.}
Mayo-2016 contains paired quarter-dose/full-dose abdominal CT images. The dataset includes 5,936 paired $512\times512$ slices with 1-mm slice thickness from ten anonymized patients. Following the common patient-level protocol, nine patients are used for training and one patient is used for testing. The quarter-dose CT images are treated as LDCT inputs, and the corresponding full-dose CT images are used as NDCT targets.

\textbf{Mayo-2020.}
Mayo-2020 provides clinical CT image/projection data together with reduced-dose data generated from clinical projection measurements. Our experimental subset contains 1,352 paired LDCT/NDCT slices from six chest studies and six abdomen studies, with 968 slices for training and 384 slices for testing. This dataset complements Mayo-2016 by covering both thoracic and abdominal anatomies.

\textbf{QIN-Lung.}
QIN-Lung is used as a projection-domain synthetic LDCT degradation benchmark. It contains 3,841 NDCT images from 46 patients, with an image size of $512\times512$ and a slice thickness of 5 mm. Since paired clinical LDCT/NDCT scans are not directly available, LDCT inputs are synthesized from NDCT images through simulated CT acquisition. For each NDCT image, TIGRE \citep{tigre} is used to perform forward projection with 360 uniformly distributed projection angles from $0$ to $2\pi$, producing projection data with 1024 detector bins per view. Poisson noise and zero-mean Gaussian noise are then injected into the projection data to simulate low-dose acquisition, and the degraded projections are reconstructed by ordered-subset simultaneous algebraic reconstruction technique (OS-SART) \citep{ossart} with 100 iterations. The original NDCT images serve as denoising targets. We use 45 patients for training and one held-out patient for testing, and all reported QIN-Lung results correspond to the $\lambda=5\times10^6$ setting.

\textbf{LoDoPaB.}
LoDoPaB is a low-dose parallel-beam CT benchmark derived from LIDC/IDRI chest CT data, providing simulated low-photon-count observations and reference reconstructions. We use it as a reconstruction-based denoising benchmark to evaluate generalization beyond Mayo-style paired clinical data. Our local pilot split contains 103 training patients and 18 testing patients, with the native $362\times362$ resolution preserved. Unlike the other datasets, LoDoPaB is processed in the normalized $[0,1]$ image domain; metrics are computed after a fixed $[-160,240]$ window-equivalent mapping, while percentile windowing is used only for visualization.

\textbf{Preprocessing and input protocol.}
For Mayo-2016, Mayo-2020, and QIN-Lung, all CT images are converted to Hounsfield units (HU), clipped to the soft-tissue window $[-160,240]$ HU, and linearly normalized to $[0,1]$:
\begin{equation}
x =
\frac{
\mathrm{clip}(x_{\mathrm{HU}}, -160, 240) + 160
}{400}.
\end{equation}
All models use the same 2.5D input protocol. Three adjacent LDCT slices are concatenated channel-wise as input, and the center NDCT slice is used as the denoising target. Mayo-2016, Mayo-2020, and QIN-Lung are processed at $512\times512$ resolution, whereas LoDoPaB keeps its native $362\times362$ resolution. For visual comparison, all methods use the same window range, crop location, and color scale within each dataset.

\textbf{Evaluation metrics.}
We employ five widely used quantitative metrics to evaluate LDCT denoising performance, including peak signal-to-noise ratio (PSNR), structural similarity index (SSIM), feature similarity index (FSIM) \citep{fsim}, noise quality metric (NQM) \citep{nqm}, and visual information fidelity (VIF) \citep{vif}. Throughout evaluation, the NDCT reference image is used as the GT.

PSNR and SSIM are used as primary pixel-level and structural similarity metrics, respectively. PSNR measures the reconstruction fidelity in terms of logarithmic mean squared error, while SSIM evaluates structural consistency by comparing local luminance, contrast, and structural correlation between the denoised  image and the reference.

FSIM evaluates image quality based on phase congruency and gradient magnitude, which are highly correlated with human visual perception. It is defined as:
\begin{equation}
\mathrm{FSIM} = \frac{\sum_{x \in \Omega} S_L(x)\cdot PC_m(x)}{\sum_{x \in \Omega} PC_m(x)},
\end{equation}
where $PC_m(x)$ denotes the phase congruency map and $S_L(x)$ represents the similarity between gradient magnitudes of the reference and denoised  images. FSIM emphasizes structural and edge preservation, making it particularly suitable for evaluating anatomical detail recovery in medical imaging.

NQM measures perceptual image quality based on a human visual system (HVS) model. It can be expressed as:
\begin{equation}
\mathrm{NQM} = \frac{\sum_{i} W_i \cdot E_i}{\sum_{i} W_i},
\end{equation}
where $E_i$ denotes local error sensitivity and $W_i$ represents perceptual weighting based on contrast sensitivity functions. NQM evaluates how well noise and blur artifacts are suppressed while maintaining perceptual fidelity.

VIF quantifies the amount of visual information preserved in the denoised  image relative to the reference based on natural scene statistics. It is defined as:
\begin{equation}
\mathrm{VIF} = \frac{I(\mathbf{x}; \mathbf{f}|\mathbf{z})}{I(\mathbf{x}; \mathbf{f})},
\end{equation}
where $I(\cdot)$ denotes mutual information between the reference image $\mathbf{x}$ and the distorted observation $\mathbf{f}$ under natural scene modeling. VIF reflects the preservation of diagnostically relevant information during denoising.

For all metrics, higher values indicate better denoising quality.

\textbf{Implementation details.}
All experiments are conducted on NVIDIA GeForce RTX 5090 GPU. The model input contains the target LDCT slice and its two spatially adjacent slices, forming a three-slice 2.5D input. The proposed model is trained with Adam using an initial learning rate of $2\times10^{-4}$, batch size 4, and 100,000 optimization iterations. The total sampling steps T were set to 1000. For the ProSAC-CT complete model, the reverse trajectory is divided into early, middle, and late stages for multi-stage diffusion training and stage-aware denoising.

\begin{figure*}[!t]
\centering
\includegraphics[width=0.8\textwidth]{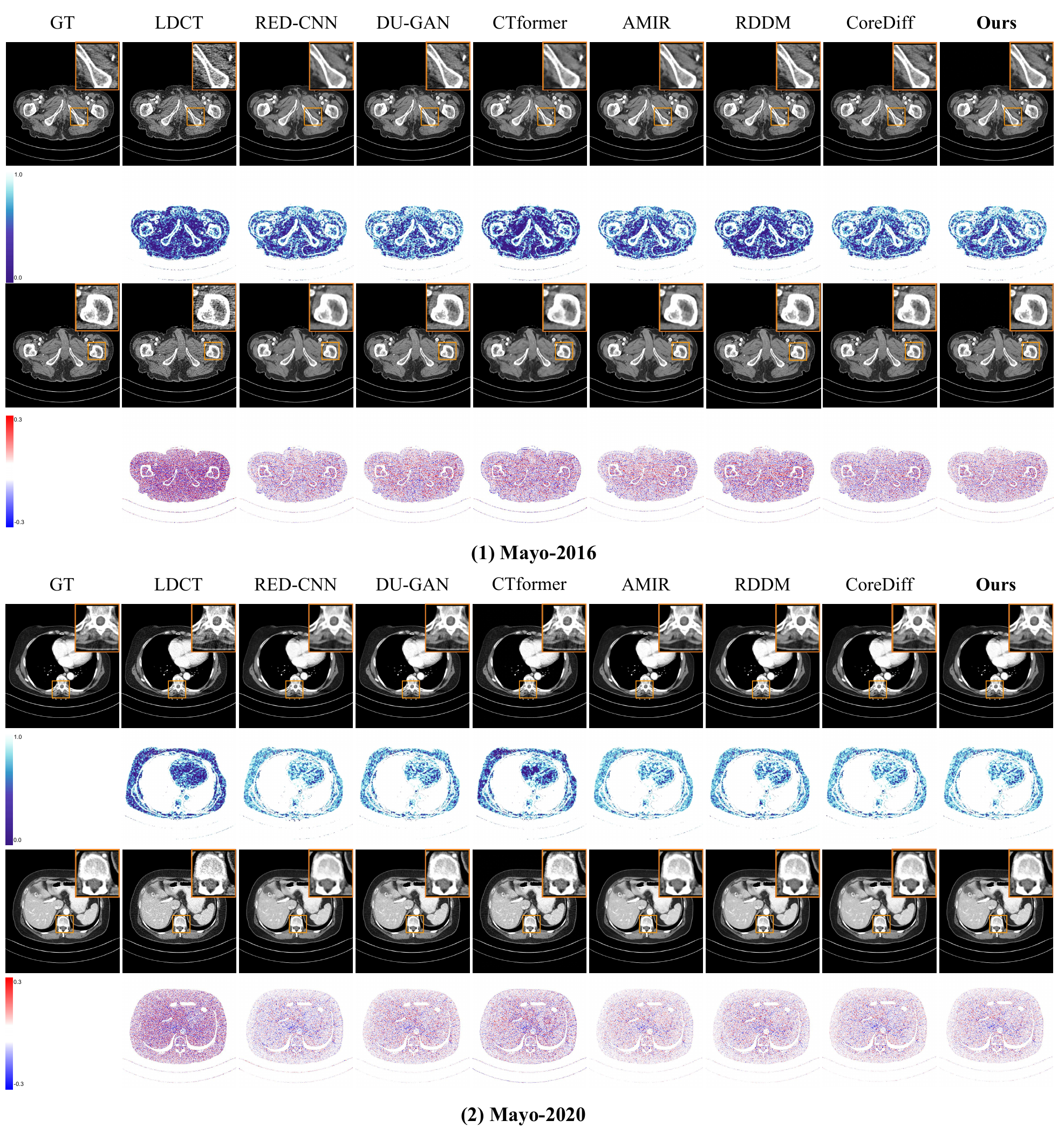}
\caption{Visualization of structural fidelity and residual error analysis on Mayo-2016 and Mayo-2020. For each dataset, two representative cases are shown with zoomed ROIs, local SSIM maps, and signed residual error maps computed with respect to GT. In the residual maps, red and blue indicate overestimation and underestimation, respectively. Both datasets use a soft-tissue CT display window of [-160, 240] HU.}
\label{fig:sota1}
\end{figure*}

\begin{figure*}[!t]
\centering
\includegraphics[width=0.8\textwidth]{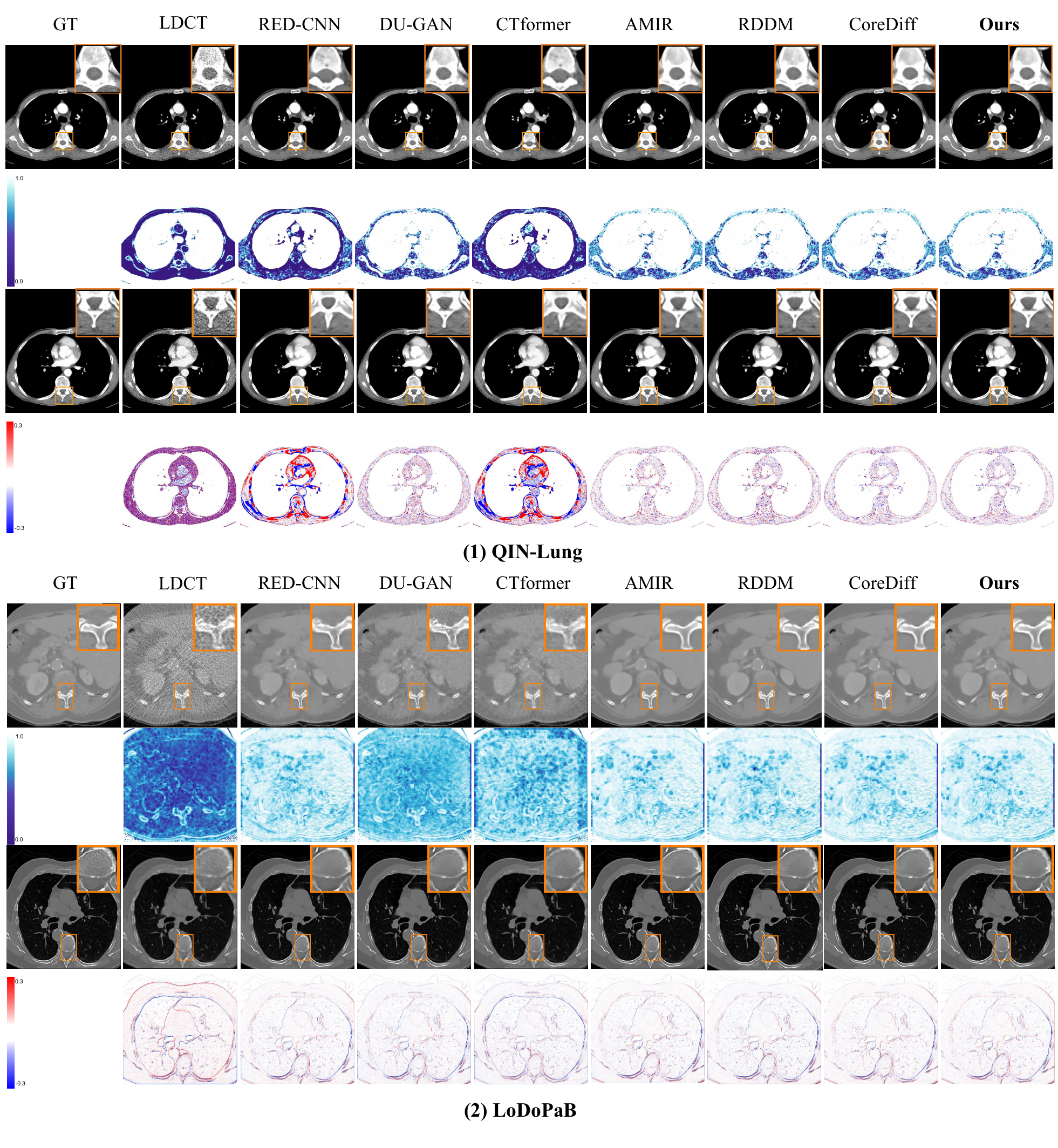}
\caption{Visualization of structural fidelity and residual error analysis on QIN-Lung and LoDoPaB. For each dataset, two representative cases are shown with zoomed ROIs, local SSIM maps, and signed residual error maps computed with respect to GT. In the residual maps, red and blue indicate overestimation and underestimation, respectively. QIN-Lung uses a soft-tissue display window of [-160, 240] HU, while LoDoPaB uses a percentile-based visualization window.}
\label{fig:sota2}
\end{figure*}

\begin{table*}[t]
\centering
\caption{Component ablation study on Mayo-2016. APGC, RFDDS, and $\mathrm{TD}^{3}$ denote the three core components of ProSAC-CT. Values are reported as mean with standard deviation. PSNR is measured in dB, SSIM is reported as a percentage, and higher values indicate better performance for all listed metrics. Best results are highlighted in bold.}
\label{tab:ct_ablation}
\scriptsize
\setlength{\tabcolsep}{6pt}
\renewcommand{\arraystretch}{1.0}
\newcommand{\stdt}[1]{{\fontsize{5}{6}\selectfont$\pm$#1}}
\begin{tabular*}{\textwidth}{@{\extracolsep{\fill}}lccc|ccccc}
\toprule
\textbf{Model} 
& \textbf{APGC} 
& \textbf{RFDDS} 
& \textbf{$\mathbf{TD}^{3}$} 
& \textbf{PSNR $\uparrow$} 
& \textbf{SSIM (\%) $\uparrow$} 
& \textbf{FSIM $\uparrow$} 
& \textbf{NQM $\uparrow$} 
& \textbf{VIF $\uparrow$} \\
\midrule
M0 (Baseline) 
&  &  &  
& 27.7347\stdt{1.8401} 
& 83.44\stdt{0.15} 
& 0.9421\stdt{0.0014} 
& 24.9625\stdt{0.1850} 
& 0.4086\stdt{0.0047} \\
M1 
& $\checkmark$ &  &  
& 29.3800\stdt{1.7621}
& 86.26\stdt{0.12} 
& 0.9486\stdt{0.0013} 
& 26.4318\stdt{0.1715} 
& 0.4472\stdt{0.0045} \\
M2 
& $\checkmark$ & $\checkmark$ &  
& 29.5400\stdt{1.7632} 
& 86.79\stdt{0.12} 
& 0.9517\stdt{0.0012} 
& 26.8754\stdt{0.1668} 
& 0.4615\stdt{0.0044} \\
M3 
& $\checkmark$ &  & $\checkmark$ 
& 30.0200\stdt{1.7322}
& 86.92\stdt{0.11} 
& 0.9543\stdt{0.0012} 
& 27.3926\stdt{0.1610} 
& 0.4748\stdt{0.0043} \\
\textbf{M4 (Ours)} 
& $\checkmark$ & $\checkmark$ & $\checkmark$ 
& \textbf{31.0847}\stdt{1.7026} 
& \textbf{88.46}\stdt{0.12} 
& \textbf{0.9587}\stdt{0.0011} 
& \textbf{28.4132}\stdt{0.1522} 
& \textbf{0.4986}\stdt{0.0042} \\
\bottomrule
\end{tabular*}
\end{table*}
\subsection{Comparison with State-of-the-Art Models}

We compare ProSAC-CT with representative CNN-, GAN-, Transformer-, and diffusion-based denoising methods, including RED-CNN, UNAD, DU-GAN, CTformer, AMIR, I2SB, ResShift, RDDM, and CoreDiff under the same preprocessing, data split, and input protocol. 

Table~\ref{tab:sota_all_ct} summarizes quantitative comparison across four CT degradation benchmarks. ProSAC-CT achieves the best performance on all datasets in terms of PSNR, SSIM, FSIM, NQM, and VIF. On the paired Mayo datasets, our method improves PSNR by 1.70 dB on Mayo-2016 and 1.78 dB on Mayo-2020 over the strongest competing approaches. On the synthetic benchmarks, consistent gains of 1.68 dB on QIN-Lung and 1.72 dB on LoDoPaB are observed, demonstrating the robustness of the proposed method under both real-world and simulated degradation settings. In addition to PSNR gains, ProSAC-CT also achieves consistent improvements in SSIM, FSIM, NQM, and VIF, suggesting that the improvement is not limited to pixel-wise error reduction but also extends to structural similarity, perceptual quality, and information fidelity.

Figures~\ref{fig:sota1} and \ref{fig:sota2} present qualitative comparisons across the four datasets using selected representative baselines for visual readability. As shown in Figs.~\ref{fig:sota1} and \ref{fig:sota2}, existing methods tend to show over-smoothed structures, residual noise, or localized intensity deviations, especially around low-contrast tissue interfaces and boundary regions. In contrast, ProSAC-CT produces clearer anatomical boundaries and more faithful intensity distributions. Similar trends are observed on Mayo-2020, QIN-Lung, and LoDoPaB, where ProSAC-CT preserves fine structural details while suppressing noise-induced distortions. The local SSIM maps and signed residual error maps further provide complementary visual evidence of structural fidelity and intensity accuracy. Compared with competing methods, ProSAC-CT shows stronger local structural consistency and smaller residual deviations in both homogeneous tissues and boundary regions.

\subsection{LDCT-Derived Anatomical Prior Analysis}
\label{sec:prior_heatmap}

\begin{figure}[H]
\centering
\includegraphics[width=0.8 \linewidth]{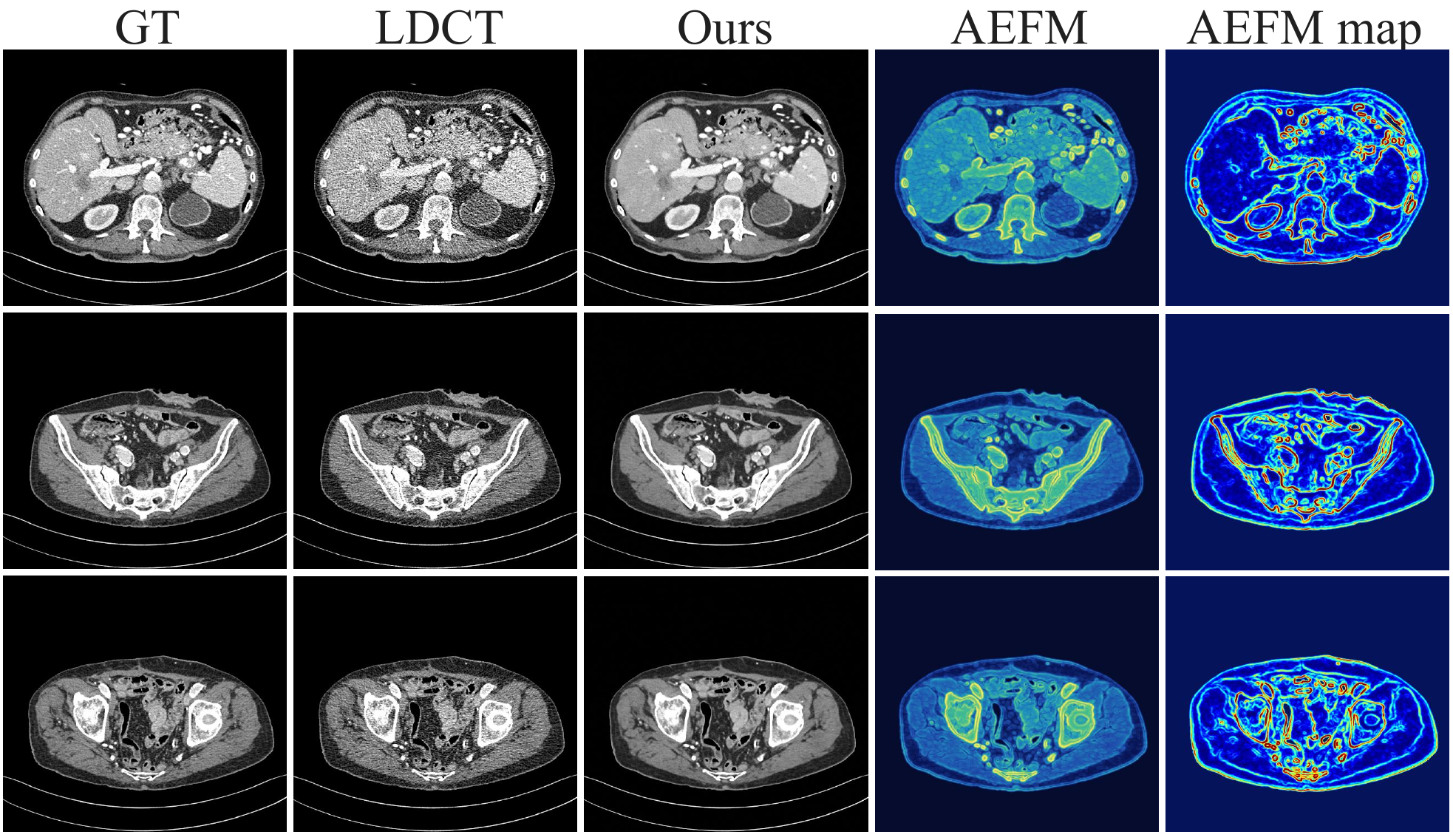}
\caption{Visualization of the LDCT-derived anatomical prior on Mayo-2016. Columns show the GT, LDCT input, ProSAC-CT output, anatomical-edge enhanced feature map (AEFM), and the heatmap of AEFM (AEFM map). CT images are displayed using the matched [-160, 240] HU window. }
\label{fig:heatmap}
\end{figure}

Figure~\ref{fig:heatmap} visualizes the LDCT-derived anatomical prior produced by the Anatomical Prior-Guided Conditioning (APGC) module. This analysis aims to examine whether useful structural information can still be extracted from noisy LDCT images. The visualization includes the GT image, LDCT input, ProSAC-CT output, anatomical-edge enhanced feature map (AEFM), and the heatmap of AEFM (AEFM map). The AEFM map is computed by spatially averaging the channel-wise activations of the AEFM representation, reflecting spatially adaptive anatomical guidance rather than uniform feature amplification across the image.

Although the LDCT inputs contain strong noise and texture degradation, the AEFM retains clear anatomical responses around organ contours, bone structures, and tissue interfaces. This is consistent with the design of APGC, where MedDINOv3 provides anatomical features while Sobel and Laplacian operators complement them with coarse contour and boundary cues. The AEFM map further shows that the prior responses are concentrated in structurally meaningful regions rather than being uniformly distributed over noisy backgrounds.
These observations support the reliability of using LDCT-derived anatomical prior for conditional diffusion denoising. The prior retains spatially aligned structural cues from the degraded input, which can guide ProSAC-CT to preserve anatomical boundaries while suppressing noise. The denoised  images are also visually closer to the GT images than the LDCT inputs, indicating that APGC provides effective structural guidance rather than simply amplifying LDCT noise. This visual behavior is consistent with the component ablation in Section \ref{sec:abl}, where adding APGC improves both structural and perceptual denoising metrics.

\subsection{Stage-Wise Spectral Analysis}

\begin{figure}[t]
\centering
\includegraphics[width=0.8\linewidth]{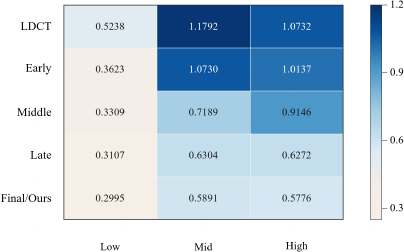}
\caption{Stage-wise spectral error trajectory of ProSAC-CT on Mayo-2016. Rows indicate the LDCT input, early-, middle-, and late-stage outputs, and the final output. Columns indicate low-, middle-, and high-frequency bands. Values denote band-wise magnitude-spectrum discrepancies with respect to GT, with lower values indicating better frequency-domain consistency. }
\label{fig:ct_stage_frequency}
\end{figure}
\begin{figure}[t]
\centering
\includegraphics[width=0.8\linewidth]{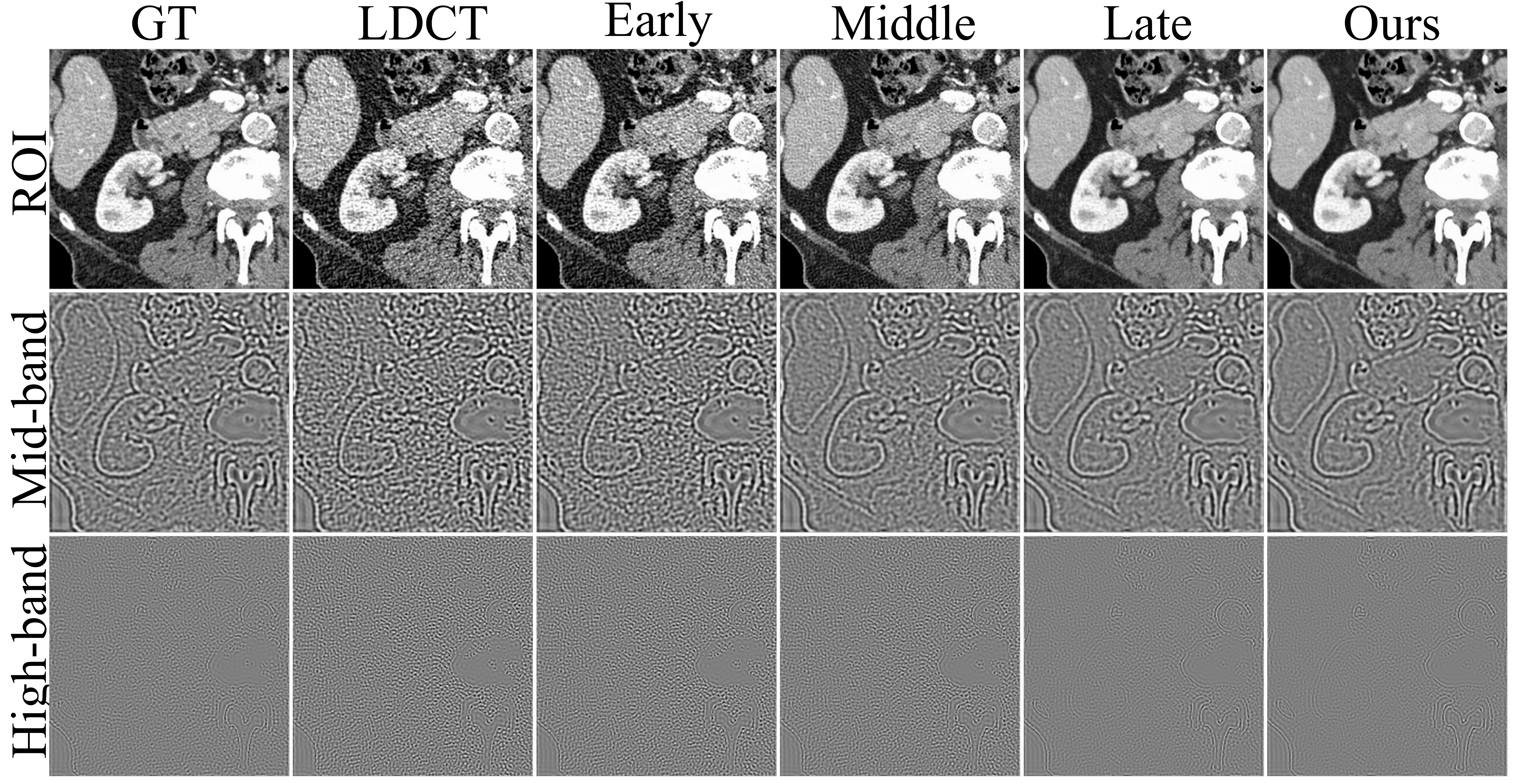}
\caption{Inverse-FFT visualization of stage-wise intermediate outputs on Mayo-2016. Columns show GT, LDCT, early-stage output, middle-stage output, late-stage output, and final output for the same ROI. Rows show the ROI image, middle-frequency component, and high-frequency component using identical row-wise display ranges.}
\label{fig:ct_stage_ifft}
\end{figure}

Figure~\ref{fig:ct_stage_frequency} presents the stage-wise spectral error trajectory of ProSAC-CT on Mayo-2016. To characterize frequency-domain \begin{figure}[!t]
\centering
\includegraphics[width=0.8\linewidth]{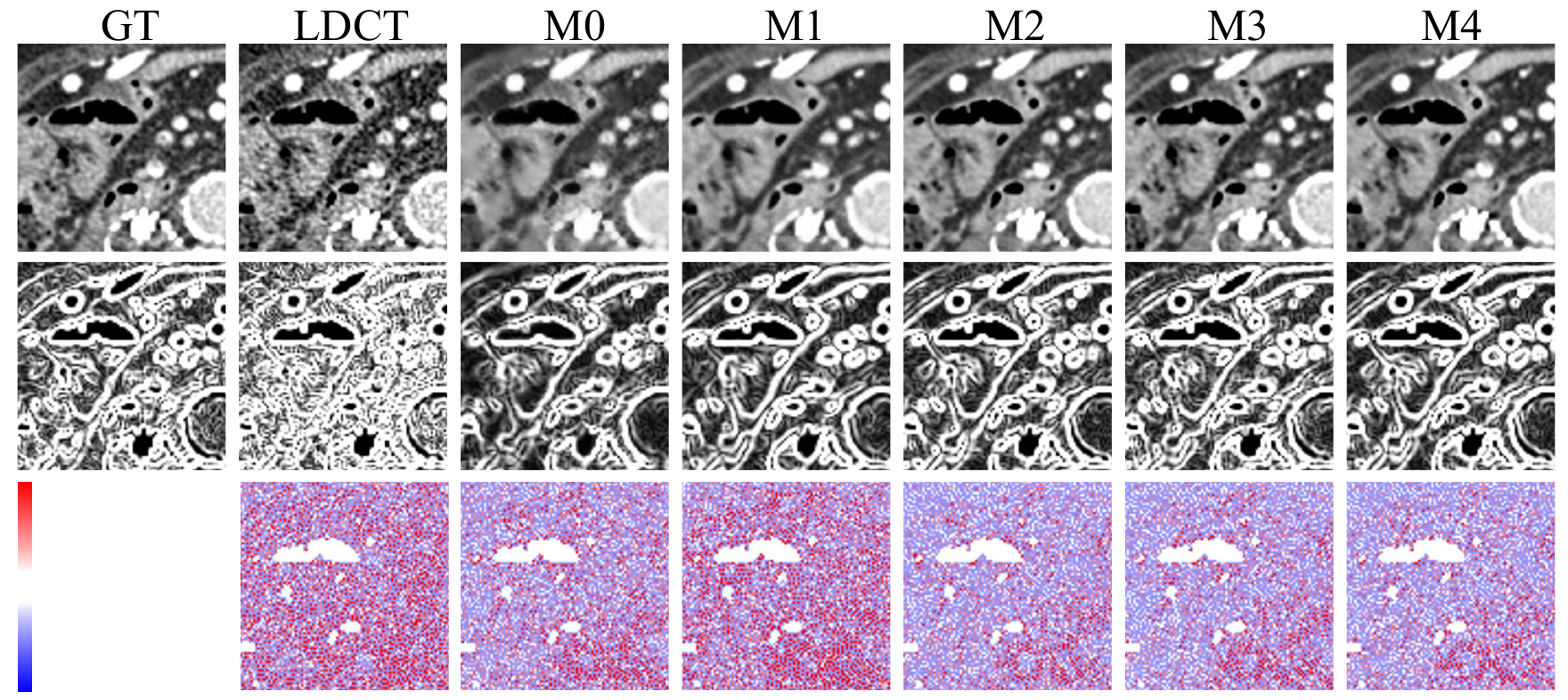}
\caption{Representative visual trend of structural recovery on Mayo-2016. Columns show GT, LDCT, M0, M1, M2, M3, and M4. From top to bottom, rows present the reconstructed ROI, structural-response map, and high-frequency residual intensity map. All variants use identical ROI coordinates and residual normalization; weaker residual responses indicate smaller high-frequency discrepancy from the GT reference.}
\label{fig:ct_visual_ablation}
\end{figure}consistency during reverse diffusion, we define band-wise spectral reconstruction errors for the saved intermediate outputs.

Let $\Omega_l$, $\Omega_m$, and $\Omega_h$ denote the low-, middle-, and high-frequency regions in the centered Fourier domain, respectively, following the same frequency partition used in RFDDS. For a given frequency band $b \in \{l, m, h\}$, the spectral error at a specific diffusion stage is computed as
\begin{equation}
E_b = \frac{1}{|\Omega_b|} \left\| \left| \mathcal{F}(\hat{x}) \right| - \left| \mathcal{F}(x) \right| \right\|_{\Omega_b,1},
\end{equation}
where $\mathcal{F}(\cdot)$ denotes the two-dimensional Fourier transform, $|\cdot|$ represents the magnitude spectrum, $\hat{x}$ is the denoised  image at a given diffusion stage, and $x$ is the NDCT reference. Thus, $E_l$, $E_m$, and $E_h$ quantify the mean magnitude-spectrum discrepancy in the low-, middle-, and high-frequency bands.

The spectral error trajectory reveals a clear stage-dependent refinement pattern during reverse diffusion. Low-frequency errors are reduced from the early stage, indicating early stabilization of global anatomical structures and intensity distributions. Middle- and high-frequency errors are progressively reduced in later stages, corresponding to boundary refinement, fine-detail consolidation, and residual noise suppression. The decrease is not a rigid one-to-one mapping between a single stage and a single frequency band. Instead, the results exhibit a progressive refinement behavior, where frequency components are gradually recovered in a coarse-to-fine manner along the diffusion trajectory.

\begin{table*}[!t]
    \centering
    \caption{Downstream six-class anatomical-region classification on Mayo-2020. LDCT input and NDCT target are included as references. For each classifier backbone, columns report F1, BAcc, and AUC. Higher values indicate better performance. Best values among denoised images are boldfaced.}
    \label{tab:ct_downstream}
    \scriptsize
    \setlength{\tabcolsep}{3pt}
    \renewcommand{\arraystretch}{1.0}
    \begin{tabular*}{\textwidth}{@{\extracolsep{\fill}}lccccccccc@{}}
        \toprule
        & \multicolumn{3}{c}{\textbf{ResNet50}} 
        & \multicolumn{3}{c}{\textbf{Swin-Tiny}} 
        & \multicolumn{3}{c}{\textbf{MambaOut-Tiny}} \\
        \cmidrule(lr){2-4} \cmidrule(lr){5-7} \cmidrule(lr){8-10}
        \textbf{Model}
        & \textbf{F1} & \textbf{BAcc} & \textbf{AUC}
        & \textbf{F1} & \textbf{BAcc} & \textbf{AUC}
        & \textbf{F1} & \textbf{BAcc} & \textbf{AUC} \\
        \midrule
        LDCT input & 0.4840 & 0.4720 & 0.8318 & 0.3854 & 0.4251 & 0.8033 & 0.5086 & 0.4949 & 0.8504 \\
        NDCT target & 0.5566 & 0.5515 & 0.8579 & 0.5088 & 0.5543 & 0.8568 & 0.5511 & 0.5629 & 0.8701 \\
        \midrule
        RED-CNN & 0.3756 & 0.4475 & 0.8117 & 0.3299 & 0.4286 & 0.8060 & 0.3508 & 0.4716 & 0.8389 \\
        UNAD & 0.3910 & 0.4559 & 0.8126 & 0.3916 & 0.4734 & 0.8386 & 0.3903 & 0.4828 & 0.8406 \\
        DU-GAN & 0.5099 & 0.5265 & 0.8431 & 0.4509 & 0.4924 & 0.8325 & 0.4803 & 0.5117 & 0.8438 \\
        CTformer & 0.4727 & 0.5013 & 0.8104 & 0.4008 & 0.4284 & 0.7904 & 0.4107 & 0.5046 & 0.8135 \\
        AMIR & 0.4168 & 0.4837 & 0.8262 & 0.3903 & 0.4927 & 0.8319 & 0.4090 & 0.4903 & 0.8384 \\
        I2SB & 0.4910 & 0.4759 & 0.8226 & 0.3916 & 0.4734 & 0.8186 & 0.3903 & 0.4828 & 0.8306 \\
        ResShift & 0.4990 & 0.5061 & 0.8159 & 0.4365 & 0.5313 & 0.8116 & 0.4787 & 0.5536 & 0.8356 \\
        RDDM & 0.4888 & 0.5111 & 0.8040 & 0.4557 & 0.5002 & 0.8173 & 0.4780 & 0.5512 & 0.8281 \\
        CoreDiff & 0.5201 & 0.5138 & 0.8244 & 0.4774 & 0.5027 & 0.8292 & 0.4960 & 0.5497 & 0.8317 \\
        \addlinespace[2pt]
        \textbf{Ours} & \textbf{0.5358} & \textbf{0.5349} & \textbf{0.8442} & \textbf{0.4875} & \textbf{0.5426} & \textbf{0.8422} & \textbf{0.5188} & \textbf{0.5611} & \textbf{0.8626} \\
        \bottomrule
    \end{tabular*}
\end{table*}

\begin{figure*}[!t]
\centering
\includegraphics[width=\textwidth]{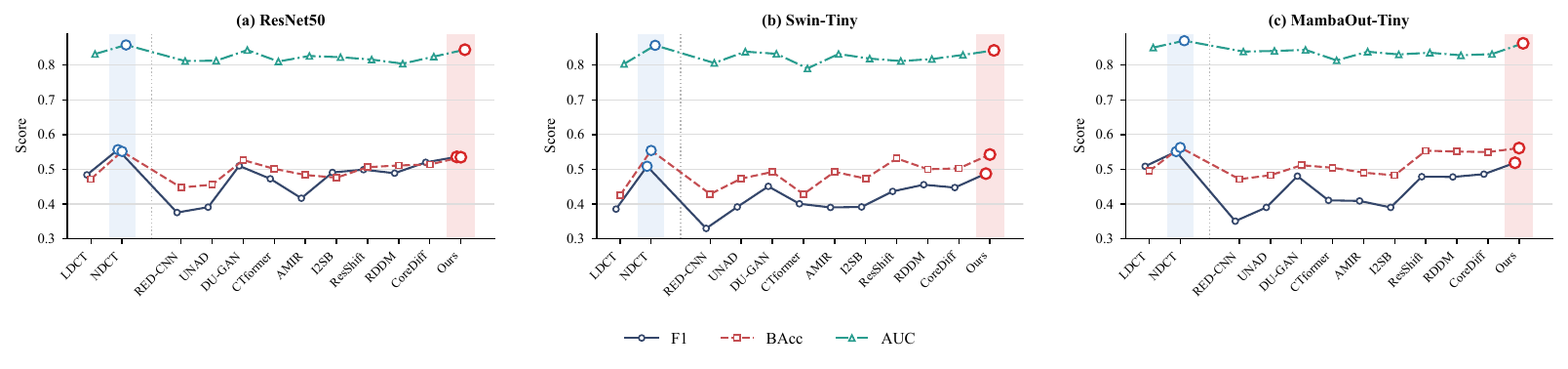}
\caption{Downstream six-class anatomical-region classification on Mayo-2020. The plots report F1, BAcc, and AUC for ResNet50, Swin-Tiny, and MambaOut-Tiny across LDCT, NDCT, competing denoised  images, and Ours. NDCT and Ours are shaded for comparison. Higher values indicate better downstream anatomical-region recognition performance.}
\label{fig:ct_downstream_radar}
\end{figure*}

Figure \ref{fig:ct_stage_ifft} provides an image-space interpretation through inverse-FFT decomposition. The ROI row shows the stage-wise image outputs. The middle-frequency row emphasizes boundary transitions and local structural interfaces. The high-frequency row contains noise-sensitive responses and fine details. Across the saved intermediate outputs, ProSAC-CT gradually moves from coarse anatomical stabilization to boundary refinement and high-frequency cleanup. This supports the motivation of using time-step-decoupling denoising with stage-specific frequency-enhanced representations.

This analysis supports the methodological motivation of ProSAC-CT by showing that its advantage is not limited to final image-quality metrics. It is also reflected in the reverse diffusion trajectory, where frequency-domain discrepancies are reduced in a structured and progressive manner.


\subsection{Ablation Study}
\label{sec:abl}


Table~\ref{tab:ct_ablation} reports the component ablation results on Mayo-2016. M0 denotes the baseline residual diffusion model implemented with the RDDM denoising formulation. M1 incorporates APGC into the baseline, while M2 and M3 further introduce RFDDS and $\mathrm{TD}^{3}$ on top of M1, respectively. M4 represents the complete ProSAC-CT model with APGC, RFDDS, and $\mathrm{TD}^{3}$.

Compared with M0, M1 consistently improves all evaluation metrics, indicating that LDCT-derived anatomical priors provide effective structural guidance for residual diffusion-based denoising. Introducing RFDDS in M2 further enhances denoising performance, suggesting that frequency-domain decoupling strengthens structure-related frequency responses and reduces noise-sensitive discrepancies. M3 also improves over M1, showing that $\mathrm{TD}^{3}$ benefits denoising by adapting the reverse process to different noise regimes.

The complete model M4 achieves the best performance across all metrics, demonstrating that APGC, RFDDS, and $\mathrm{TD}^{3}$ provide complementary and cumulative contributions. Specifically, anatomical guidance improves structural consistency, frequency decoupling enhances spectral denoising, and time-step-aware decoding stabilizes progressive denoising.

Figure~\ref{fig:ct_visual_ablation} provides a visual comparison of the ablation variants. M1 shows clearer boundary preservation after introducing APGC, M2 reduces high-frequency discrepancy with RFDDS, and M3 produces more coherent structural responses through $\mathrm{TD}^{3}$-based stage-aware denoising. The complete model M4 yields the most faithful ROI reconstruction, the clearest structural response, and the weakest residual response, further supporting the effectiveness of the three proposed components.

\subsection{Downstream Anatomical Classification}

To assess the downstream utility of denoised images for medical image analysis beyond pixel-level fidelity, we perform six-class anatomical-region classification on Mayo-2020 using three independent classifier backbones, including ResNet50~\citep{he2015deepresiduallearningimage}, Swin-Tiny~\citep{liu2021swintransformerhierarchicalvision}, and MambaOut-Tiny~\citep{yu2024mambaoutreallyneedmamba}. Labels are defined by combining the scan-level body region, abdomen or chest, with the patient-wise axial tertile computed from the slice position within each scan, yielding ABDOMEN\_Z1--ABDOMEN\_Z3 and CHEST\_Z1--CHEST\_Z3.
 F1, balanced accuracy (BAcc), and AUC are used to evaluate class-balanced recognition performance and ranking ability. Specifically, F1 measures class-balanced precision--recall trade-offs, BAcc evaluates average sensitivity across classes, and AUC reflects the ranking ability of the classifier under multi-class settings.

As shown in Table~\ref{tab:ct_downstream}, ProSAC-CT achieves the best performance among denoised  images across all three classifier backbones and all evaluation metrics. Compared with the strongest denoised image methods for each metric, ProSAC-CT improves F1, BAcc, and AUC by 0.0157, 0.0084, 0.0011 with ResNet50, by 0.0101, 0.0113, 0.0036 with Swin-Tiny, and by 0.0228, 0.0075, 0.0188 with MambaOut-Tiny. These gains indicate that ProSAC-CT better preserves anatomy-discriminative information than competing denoised image baselines.

Beyond the comparison among denoised images, we further examine ProSAC-CT against the LDCT input and the NDCT target. Compared with the LDCT input, ProSAC-CT improves all three downstream metrics across all classifier backbones, with largest absolute gains of 0.1021 in F1, 0.1175 in BAcc, and 0.0389 in AUC. More importantly, ProSAC-CT achieves the closest denoised-image performance to the NDCT target across the reported metrics. Notably, several denoised image methods underperform the LDCT input in downstream classification, suggesting that visual denoising does not necessarily preserve recognition-relevant anatomical cues. These results show that ProSAC-CT outputs can effectively support downstream clinical analysis under low-dose acquisition requirements.

Figure~\ref{fig:ct_downstream_radar} provides a visual summary of the downstream classification results across the three classifier backbones. ProSAC-CT consistently shows the strongest denoised image profile, further supporting its downstream utility across different recognition architectures.

\FloatBarrier

\section{Conclusion}

We presented ProSAC-CT, a progressive spectral--anatomical co-guided multi-stage diffusion model for image-domain LDCT denoising. By integrating LDCT-derived anatomical-prior-guided conditioning, residual frequency-domain decoupling, and time-step-decoupling denoising, ProSAC-CT improves anatomy-aware representation learning, frequency-sensitive denoising, and stage-aware reverse diffusion. Experiments on four LDCT degradation benchmarks demonstrate consistent improvements over representative methods in image fidelity, structural similarity, perceptual quality, and information preservation. Stage-wise spectral analysis and ablation studies further verify the progressive denoising behavior and the complementary contributions of the proposed components.

Beyond these quantitative gains, downstream anatomical-region classification on Mayo-2020 shows that ProSAC-CT better preserves task-relevant anatomical information than competing denoised images. This suggests that its denoising process suppresses LDCT noise while retaining the anatomical structures needed for subsequent analysis. The downstream classification results further show that ProSAC-CT narrows the performance gap between LDCT and NDCT, indicating that the denoised images approach the downstream task performance of NDCT. Therefore, ProSAC-CT provides a practical image-domain denoising solution for low-dose CT applications that aim to reduce patient radiation exposure, including low-dose screening, follow-up imaging, and downstream computer-aided medical image analysis.